\documentclass[journal]{IEEEtran}
\usepackage{amsmath}
\usepackage{graphicx}
\usepackage{threeparttable}
\usepackage{multirow}
\usepackage{colortbl}
\usepackage{color}
\usepackage{fancyvrb}

\usepackage{url}
\usepackage{caption}
\usepackage{subfig}

\ifCLASSINFOpdf
\else
\fi
\begin{document}
%
\title{Unsupervised End-to-end Learning \\for Deformable Medical Image Registration}
%
%
%

\author{Siyuan Shan, Wen Yan, Xiaoqing Guo, Eric I-Chao Chang, Yubo Fan and Yan Xu*
\thanks{This work is supported by Microsoft Research under the eHealth program, the National Natural Science Foundation in China under Grant 81771910, the National Science and Technology Major Project of the Ministry of Science and Technology in China under Grant 2017YFC0110903, the Beijing Natural Science Foundation in China under Grant 4152033, the Technology and Innovation Commission of Shenzhen in China under Grant shenfagai2016-627, Beijing Young Talent Project in China, the Fundamental Research Funds for the Central Universities of China under Grant SKLSDE-2017ZX-08 from the State Key Laboratory of Software Development Environment in Beihang University in China, the 111 Project in China under Grant B13003. \emph{Asterisk indicates corresponding author.}}
\thanks{Siyuan Shan, Wen Yan, Xiaoqing Guo, Yubo Fan and Yan Xu are with State Key Laboratory of Software Development Environment and Key Laboratory of Biomechanics and Mechanobiology of Ministry of Education and Research Institute of Beihang University in Shenzhen, Beihang University, Beijing 100191, China (email: shansiliu@outlook.com; yanwen@buaa.edu.cn; guoxiaoqing@buaa.edu.cn; yubofan@buaa.edu.cn; xuyan04@gmail.com).}
\thanks{Eric I-Chao Chang, and Yan Xu are with Microsoft Research, Beijing 100080, China (email:echang@microsoft.com; xuyan04@gmail.com).}
}


%
%

\markboth{Journal of \LaTeX\ Class Files,~Vol.~14, No.~8, August~2015}%
{Shell \MakeLowercase{\textit{et al.}}: Bare Demo of IEEEtran.cls for IEEE Journals}
%



\maketitle

\begin{abstract}
We propose a registration algorithm for 2D CT/MRI medical images with a new unsupervised end-to-end strategy using convolutional neural networks. The contributions of our algorithm are threefold: (1) We transplant traditional image registration algorithms to an end-to-end convolutional neural network framework, while maintaining the unsupervised nature of image registration problems. The image-to-image integrated framework can simultaneously learn both image features and transformation matrix for registration. (2) Training with additional data without any label can further improve the registration performance by approximately 10\%. (3) The registration speed is 100x faster than traditional methods. The proposed network is easy to implement and can be trained efficiently. Experiments demonstrate that our system achieves state-of-the-art results on 2D brain registration and achieves comparable results on 2D liver registration. It can be extended to register other organs beyond liver and brain such as kidney, lung, and heart.
\end{abstract}


%
\IEEEpeerreviewmaketitle

\section{Introduction}

\begin{figure}[!tp]
\centering
\subfloat[Fixed]{\includegraphics[width=0.2\linewidth]{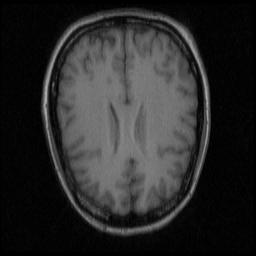}}
\hspace{0.02\linewidth}
\subfloat[Moving]{\includegraphics[width=0.2\linewidth]{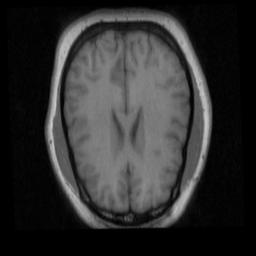}}
\hspace{0.02\linewidth}
\subfloat[Deformation]{\includegraphics[width=0.2\linewidth]{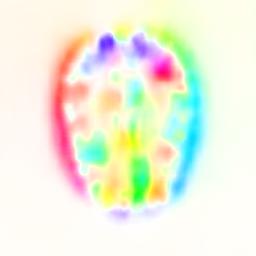}}
\hspace{0.02\linewidth}
\subfloat[Warped]{\includegraphics[width=0.2\linewidth]{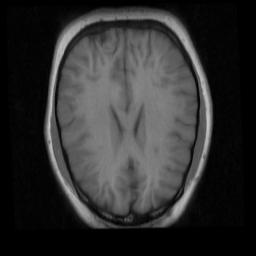}}
\hspace{0.02\linewidth}
\caption{Illustration of image registration. Given a fixed image (a) and a moving image (b), a deformation field (c) is predicted to warp the moving image so that (d) and (a) are spatially aligned. Note that the deformation field is color coded according to .}
\label{fig:intro}
\end{figure}

\IEEEPARstart{M}{edical} image registration plays an important role in medical image processing and analysis. As far as brain registration is concerned, accurate alignment of the brain boundary and corresponding structures inside the brain such as hippocampus is crucial for monitoring brain cancer development. As illustrated in Figure \ref{fig:intro}, image registration refers to the process of revealing the spatial correspondence between two images. Several image registration toolkits such as ITK \cite{ibanez2005itk}, ANTs \cite{avants2010ants} and Elastix \cite{klein2010elastix} have been developed to facilitate research reproduction.

A wide variety of medical registration algorithms have been developed in the past \cite{ashburner2007fast,vercauteren2009diffeomorphic,song2010lung,klein2010elastix,thirion1998image,xu20163d}, focusing primarily on unsupervised methods. These algorithms select a transformation model, define a metric that measures the similarity of two images to be registered, and iteratively update the transformation parameters or deformation field to optimize the defined metric. A fraction of registration algorithms is learning-based \cite{guetter2005learning}. For learning-based approaches: (1) informative feature representations are difficult to obtain directly from learning and optimizing morphing or similarity function; (2) unlike image classification and segmentation, registration labels are difficult to collect. These two reasons limit the development of learning-based registration algorithms.

\begin{figure*}[!tp]
\centering
\includegraphics[width=150mm]{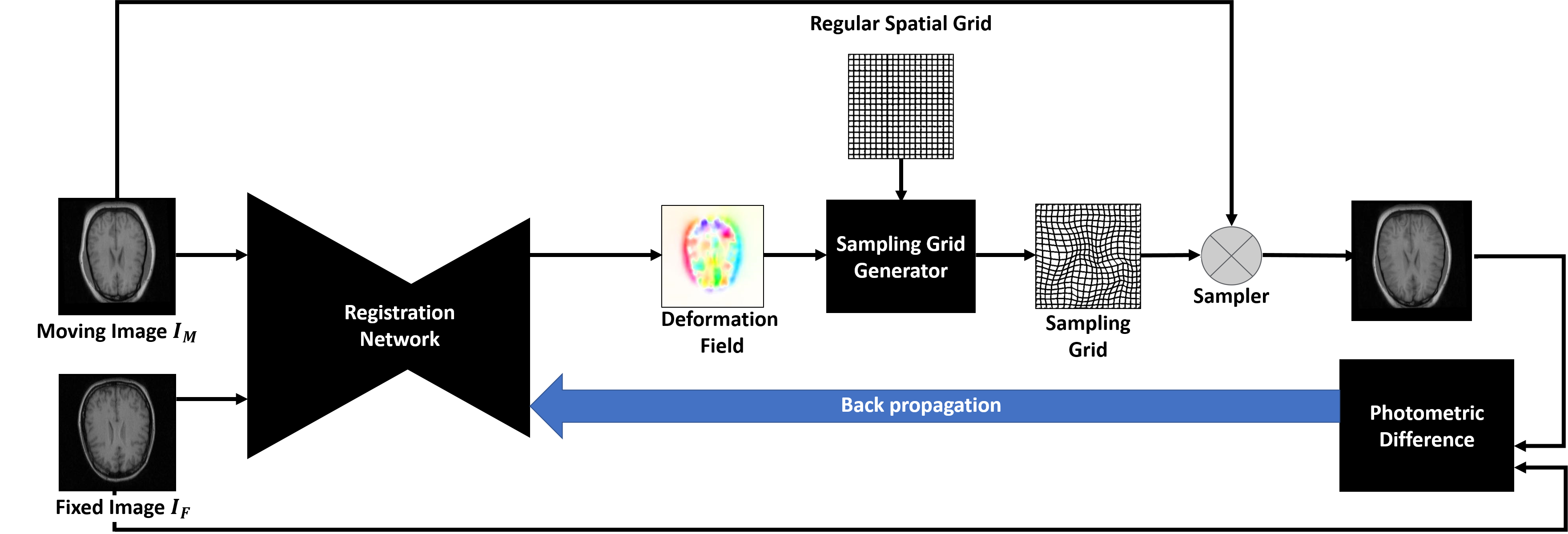}
\caption{Illustration of the unsupervised training strategy of our fully convolutional image-to-image registration network. The registration network takes two images and outputs a deformation field, which is used to produce the sampling grid. The moving image is then warped by the sampling grid via bilinear interpolation. The loss function is defined as the photometric difference between the warped image and the fixed image. The registration error can be efficiently back propagated to update the learnable parameters of the registration network for end-to-end training.}
\label{fig:framework}
\end{figure*}

Recently, the field of computer vision has witnessed a tremendous advancement triggered by deep learning technologies like convolutional neural networks (CNNs). CNNs have proven their mettle in handling image classification \cite{krizhevsky2012imagenet,simonyan2014very}, object detection \cite{ren2015faster} as well as pixel-wise prediction tasks like semantic segmentation \cite{long2015fully} and edge detection \cite{xie2015holistically}. Apart from these tasks where only a single image is processed, CNNs also have the capacity to tackle image matching and registration problems. For example, Zbontar and LeCun \cite{zbontar2016stereo} train a CNN to predict the similarity of two image patches for subsequent stereo matching. Wei et al. \cite{wei2016dense} utilize CNNs as a feature extraction tower and compute dense human body correspondence according to the feature vectors extracted. These works qualify CNNs as a potential tool for medical image registration. 

There have been few works to use CNNs for medical image registration \cite{litjens2017survey}. Yang et al. \cite{yang2016fast} design a deep encoder-decoder network to initialize the momentum of the large deformation diffeomorphic metric mapping (LDDMM) registration model. However, their method is a patch-based algorithm and thus requires postprocessing that cannot be handled inside CNNs. Wu et al. \cite{wu2013unsupervised} adopt unsupervised deep learning to obtain features for image registration. Though good performance is achieved, their method is also a patch-based learning system and relies on other feature-based registration methods to perform image registration. Miao et al. \cite{miao2016cnn} adopt CNN regressors to directly predict transformation parameters for 2D/3D images and achieve higher registration success rates than traditional methods. But their model is not trained end-to-end and cannot perform deformable registration. Compared with patch-based training systems, image-to-image prediction can be performed by fully convolutional neural networks (FCNs) \cite{long2015fully} where pixel-wise features are predicted. Therefore, a CNN model that can perform image-to-image deformable registration through end-to-end FCNs is desired.

FlowNet \cite{fischer2015flownet} is an appropriate CNN that can directly predict optical flow from two input images using end-to-end fully convolutional networks (FCNs) with competitive accuracy at frame rates of 5 to 10 fps. FlowNet is trained on a synthetic dataset in a supervised manner, where natural image pairs with ground-truth registration parameters are generated via computer graphic techniques. However, unlike natural images, realistic medical images are difficult to generate. Consequently, the learning-based methods have not been widely used to solve medical image registration problems \cite{litjens2017survey}. To this point, it is highly desired to develop an unsupervised learning framework with end-to-end CNNs for medical image registration, which implicitly learns to predict registration parameters or deformation without ground-truth supervision.

The spatial transformer network (STN) proposed by Jaderberg et al. \cite{jaderberg2015spatial} enables neural networks to spatially transform feature maps. The process of STN is as follows: STN first generates a sampling grid according to the transformation parameters produced by neural networks. An input image can be spatially warped by the sampling grid. The warping process is implemented by bilinear interpolation, which makes STN fully-differentiable.  Several STN-based approaches have been proposed to address similar problems in natural scenes, such as optical flow estimation \cite{ren2017unsupervised, yu2016back} depth estimation \cite{garg2016unsupervised,godard2016unsupervised} and single-view reconstruction \cite{kanazawa2016warpnet}. Inspired by the recent success of STN \cite{jaderberg2015spatial},  we develop an unsupervised learning framework by combining the spatial transformer with fully convolutional neural networks for 2D medical image registration. The integrated framework can simultaneously learn both image features and transformation matrix for registration. We define the pixel-wise difference between the warped moving image and the fixed image as the loss function, the registration error can be effectively backpropagating to CNNs for learning the optimal transformation parameters that minimize the registration error. As shown in Figure \ref{fig:framework}, this training strategy is very similar to the mechanism of traditional registration algorithms where no ground-truth deformation is required.

In this paper, we build an end-to-end unsupervised learning system with fully convolutional neural networks in which image-to-image medical image registration is performed as illustrated in Figure \ref{fig:framework}. Compared with FlowNet, the algorithm does not require a synthetic dataset for supervised learning. Compared with STN, our method can perform image registration in a deformation field form while STN can only perform classification; our method is for template alignment while STN is for class alignment. Besides, as an unsupervised learning model, its registration performance can be easily improved by introducing additional training data without any label.

To summarize, in this paper we develop unsupervised convolutional neural networks for 2D tissue registration via direct deformation field prediction. The contributions of our algorithm are threefold: (1) Our algorithm is an end-to-end CNN-based learning system under an unsupervised learning setting that performs image-to-image registration. (2) Training with additional data without any label can further improve the registration performance. (3) We achieve a 100x speed-up compared to traditional image registration methods. 

\section{Related Work}
Here we first describe research directly related. Then, the key components of the traditional algorithms are summarized and several works that tackle image registration problems with CNN are outlined.
\subsection{Directly Related Works}
Three existing approaches that are closely related to our work are discussed below.

Dosovitskiy et al. \cite{dosovitskiy2015flownet} propose an end-to-end fully convolutional neural net FlowNet for optical flow estimation in real time. FlowNet has an encoder-decoder architecture with skip connections. It predicts optical flow at multiple scales and each scale is predicted based on the previous scale. Compared with the nature of supervised learning of Flownet, an unsupervised architecture is utilized in this work to predict deformation field that aligns two images. 

Jaderberg et al. \cite{jaderberg2015spatial} propose the spatial transformer networks (STN) which focuses on class alignment. It shows that spatial transformation parameters (e.g. affine transformation parameters, B-Spline transformation parameters, deformation field, etc) can be implicitly learned without ground-truth supervision by optimizing a specific loss function \cite{jaderberg2015spatial}. STN is a fully differentiable module that can be inserted into existing CNNs, which makes it possible to cast the image registration task as an image reconstruction problem. Few papers focus on the registration task using STN. In this paper, we use the STN to make registration alignment in the medical  image field.

\subsection{Traditional Medical Image Registration Algorithms}
A variety of traditional medical image registration algorithms have been proposed over the past few decades \cite{ashburner2007fast,vercauteren2009diffeomorphic,song2010lung,klein2010elastix,thirion1998image}.
A successful image registration application requires several components that are correctly combined, namely the definition of the cost function, the multiresolution strategy, and the coordinate transformation model.

Cost functions, also called similarity metrics, measure how well two images are matched after transformation. The cost function is one of the most crucial parts of a registration algorithm. It is selected with regards to the types of objects to be registered. Commonly adopted cost functions are the mean squared difference \cite{kybic2003fast}, mutual information \cite{viola1997alignment}, normalized mutual information \cite{rohlfing2003volume} and normalized correlation \cite{penney1998comparison}. A regularization term is often required to penalize undesired deformations \cite{staring2007rigidity}.

The multiresolution strategy \cite{lester1999survey} is a widely adopted technique to increase registration speed and improve the stability of the optimization. A sequence of reduced resolution versions of input images is created, which forms a pyramid representation. Then registration is performed at each level of the pyramid from coarse to fine resolution consecutively, with the initial transformation of the next level being the resulting transformation of the previous level.

Coordinate transformation models are determined according to the complexity of deformations that need to be recovered. Though in some cases parametric transformation models (such as rigid, affine and B-Splines transformation) are enough to recover the underlying deformations \cite{viola1997alignment,rueckert1999nonrigid}, a more flexible non-parametric transformation model allowing for arbitrary local deformations is usually needed \cite{ashburner2007fast,vercauteren2009diffeomorphic}. Non-parametric registration aims to find a dense deformation field where each pixel is individually displaced to get a reasonable alignment of the images. In this work, we only consider non-parametric transformation models.

To date, traditional registration algorithms have achieved satisfactory performance on various datasets. However, they have a non-negligible drawback. For each pair of unseen images to be registered, traditional registration methods iteratively optimize the cost function from scratch, which seriously limits the registration speed and totally neglects the inherent registration patterns shared across images from the same dataset. In this work, we propose a fully convolutional and image-to-image registration framework to overcome the above mentioned drawbacks while maintaining competitive registration performance. It is also shown that the three components of classical methods can be easily transplanted to existing CNN frameworks.

\subsection{Supervised Learning Methods}
There have been few works to use supervised CNNs in a patch-based manner for medical image registration. Yang et al. \cite{yang2016fast} design a deep encoder-decoder network to initialize the momentum of the large deformation diffeomorphic metric mapping registration model. Sokooti et al. \cite{Sokooti:2017} train a 3D CNN to register chest CT data using artificially generated displacement vector field. Their method is also patch-based.

Compared with patch-based training systems, image-to-image prediction can be performed by fully convolutional neural networks (FCNs) \cite{long2015fully} where pixel-wise features are predicted.  Fischer et al. \cite{dosovitskiy2015flownet} propose a novel CNN model for optical flow prediction. This model is trained end-to-end on a synthetic dataset and can perform image-to-image optical flow prediction. 



Though all of these works achieve competitive performance, they are trained on synthetic datasets \cite{Sokooti:2017,dosovitskiy2015flownet} or datasets using the results of classical methods as ground truth \cite{yang2016fast}. 


\subsection{Unsupervised Learning Methods}

To obviate the need to collect real data with abundant and reliable ground-truth annotations, unsupervised learning methods become prevalent. Wu et al. \cite{wu2013unsupervised} adopt unsupervised deep learning to obtain features for image registration. Though good performance is achieved, their method is a patch-based learning system and relies on other feature-based registration methods to perform image registration. Ren et al. \cite{ren2017unsupervised} and Yu et al. \cite{yu2016back} use the spatial transformer networks (STN) \cite{jaderberg2015spatial} and optical flow produced by a CNN to warp one frame to match its previous frame. The difference between two frames after warping is used as the loss function to optimize the parameters of CNN. Their unsupervised methods do not require any ground-truth optical flow. Similarly, Garg et al. \cite{garg2016unsupervised} use an image reconstruction loss to train a network for monocular depth estimation. This work is further ameliorated by incorporating a fully differentiable training loss and left-right consistency check \cite{godard2016unsupervised}.
We follow the idea of these works to train a model for image-to-image registration in an unsupervised manner. An auxiliary loss function is also introduced to regularize the deformation field as will be discussed in the next section.

\section{Methods}
\begin{figure}[!hbtp]
\centering
\includegraphics[width=90mm]{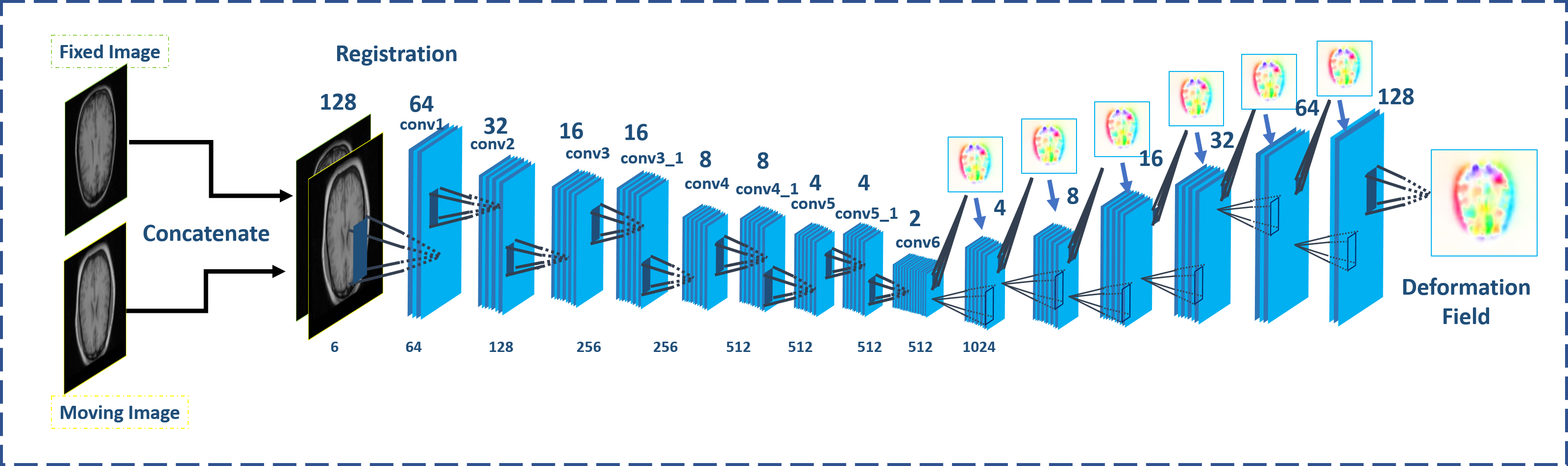}
\caption{Illustration of the detailed structures of our image registration network. The size
and number of channels of each feature map are shown. The network takes two concatenated images to be registered as input and predicts the deformation field at 7 different scales during training. During testing, only the deformation field at the largest scale is used. The skip architecture of FlowNet is not shown for simplicity.}
\label{fig:flowchart}
\end{figure}

This section introduces the problem of image registration and describes our image registration network. We introduce a novel training loss for the problem that does not require supervision in the form of ground truth deformation. Figure \ref{fig:flowchart} illustrates our neural networks for medical image registration.

\subsection{Problem Statement}
In image registration, one image $I_{M}$ called the moving image is deformed to match another image $I_{F}$ called the fixed image according to a two-dimensional dense deformation field $\mathbf{u}$. The deformed image $\tilde{I}$ is expressed as 
\begin{equation}
  \tilde{I}\left ( \mathbf{x} \right ) = I_{M}\left ( \mathbf{x}+\mathbf{u(x)} \right ) ,
\end{equation}
where $\mathbf{x}$ denotes a two-dimensional location. In this work, we attempt to estimate the optimal deformation field $\mathbf{u(x)}$ for accurate image registration.

\subsection{Unsupervised Image Registration Network}
\label{loss}
A fully convolutional network is adapted to model the complex non-linear transformation from two input images to a deformation field that aligns the input images. The deformation field prediction module is inspired by FlowNet \cite{dosovitskiy2015flownet}. FlowNet is a fully convolutional neural net originally proposed to solve the optical flow estimation problem. It takes two input images and outputs a dense optical flow/deformation field that aligns the two input images. FlowNet consists of a contracting part to capture context and an expanding part of intelligent flow field refinement. Skip connections are also included to combine high-level and low-level features. FlowNet predicts optical flow at multiple scales and each scale is predicted based on the previous scale. This design resembles the multiresolution strategy \cite{lester1999survey} adopted by traditional registration algorithms and improves the robustness of deformation field prediction. 
We adopt FlowNetSimple architecture \cite{dosovitskiy2015flownet} with one modification.
The output optical flow field of FlowNet is 4 times smaller than the input. To obtain a dense deformation field that has the same resolution as the input image, we repeat the upsampling block of FlowNet twice. The detailed structure of the proposed image registration network is shown in Figure \ref{fig:flowchart}. 

FlowNet \cite{dosovitskiy2015flownet} is originally trained in a supervised manner by minimizing the endpoint error (EPE) between the predicted flow vector and the ground truth flow averaged over all pixels. For image registration problems, however, ground truth deformation is difficult to collect. Though we can alternatively regard the result of traditional registration algorithms as ground truth to train the network, they are not universal solutions considering the variability of medical images and the inaccuracy of traditional algorithms.

The problem mentioned above is mainly caused by the inherent incongruity between the unsupervised nature of the image registration problem and the supervised training strategy of CNN. Concretely, traditional registration algorithms are not learning-based and thus unsupervised. A similarity metric that forces two images to appear similar to each other is optimized and there are no learnable parameters in traditional algorithms. The philosophy of CNN, however, is utterly different. CNN is a high-capacity learning model containing millions of learnable parameters. It is usually trained in a supervised manner where ground-truth class labels or segmentation masks are provided. Therefore, a modification of the classical CNN architecture is required to transplant traditional registration algorithms to the deep learning framework. 

In this work, the spatial transformer network (STN) \cite{jaderberg2015spatial} is inserted to our image registration network for unsupervised learning. STN is selected for two reasons. First, as it can spatially warp feature maps or images inside neural networks, the warped moving image can be successfully produced to constitute the photometric loss function. Second, its fully-differentiable property makes it possible to train the registration network end-to-end. STN contains a regular spatial grid generator and a sampler. The deformation field predicted by our image registration network is used to transform the regular spatial grid into a sampling grid. Then, the sampler uses the sampling grid to warp the input image. Bilinear interpolation is adopted during the sampling, which makes STN fully differentiable for backpropagation. By defining the pixel-wise differences (photometric differences) between the warped moving image and the fixed image as the loss function, the image registration problem becomes an image reconstruction problem. The loss functions of our image registration network are defined as follows.

\noindent\textbf{Photometric Difference Loss} We define a photometric loss $\mathcal{L}_{photometric}^{s}$ at each output scale $s$ of FlowNet and let $\mathbf{u^{s}(x)}$ denote the predicted deformation field at output scale $s$. $\mathcal{L}_{photometric}^{s}$ is $L1$ photometric image reconstruction error defined as
\begin{equation}
\mathcal{L}_{photometric}^{s} =  \sum_{\mathbf{x}\in \Omega}||\tilde{I}^{s}\left ( \mathbf{x} \right ) - I_{F}^{s}(\mathbf{x})|| ,
\end{equation}
where $\tilde{I}^{s}\left ( \mathbf{x} \right )= I_{M}^{s}\left ( \mathbf{x}+\mathbf{u^{s}(x)} \right )$ is the moving image resized to scale $s$ and warped by STN according to deformation field $\mathbf{u^{s}(x)}$, $I_{F}^{s}(\mathbf{x})$ is the fixed image resized to scale $s$ and $\Omega$ is the two-dimensional image plane. This loss function encourages the warped image to appear similar to the fixed image.

\noindent\textbf{Deformation Field Smoothness Loss}
A regularization term $\mathcal{L}_{smooth}$ is generally needed to encourage the estimated deformation field to be locally smooth. In this work two types of regularization terms $\mathcal{L}_{smoothN}$ and $\mathcal{L}_{smoothE}$ are compared. $\mathcal{L}_{smoothN}$ is a normal $L1$ penalty on the deformation field gradient $\partial \mathbf{u^{s}(x)}$,
\begin{equation}
\mathcal{L}_{smoothN}^{s} = \sum_{\mathbf{x}\in \Omega} |\partial _{x}\mathbf{u^{s}(x)}|+|\partial _{y}\mathbf{u^{s}(x)}|,
\end{equation}  
where $\partial _{x}$ and $\partial _{y}$ respectively denote partial derivatives along horizontal and vertical directions.

$\mathcal{L}_{smoothE}$ is the $L1$ penalty  weighted by an edge-aware term \cite{heise2013pm} as deformation field discontinuities often occur at image gradients,
\begin{equation}
\mathcal{L}_{smoothE}^{s} = \sum_{\mathbf{x}\in \Omega} |\partial _{x}\mathbf{u^{s}(x)}|e^{-||\partial_{x}I_{F}^{s}(\mathbf{x})||}+|\partial _{y}\mathbf{u^{s}(x)}|e^{-||\partial_{y}I_{F}^{s}(\mathbf{x})||}.
\end{equation}

\noindent\textbf{Total Loss}
The total loss $\mathcal{L}$ is the weighted sum of the above defined losses,
\begin{equation}
\mathcal{L} = \sum_{s=1}^{7}\alpha _{s} \mathcal{L}_{photometric}^{s} + \beta _{s} \mathcal{L}_{smooth}^{s} ,
\end{equation}
where there are seven output scales of the proposed deformation field prediction module.

Our framework is an end-to-end learning system and allows for fast and accurate deformation field prediction. In our experiment, we show that the proposed registration network performs well compared to traditional methods both in terms of accuracy and speed.


\section{Experiment}
We evaluate the proposed algorithm with extensive experiments on two datasets with the ground truth of the corresponding landmarks and segmentation boundaries. These two datasets respectively contain MRI brain images and CT liver images. The proposed unsupervised registration algorithm is compared to traditional registration algorithms provided by off-the-shelf toolboxes like Advanced Normalization Tools (ANTs) \cite{avants2009advanced}, Elastix \cite{klein2010elastix} and Insight Segmentation and Registration Toolkit (ITK) \cite{ibanez2005itk}. We also report the results of our baseline methods (supervised) where the outputs of the above mentioned traditional algorithms are regarded as ground truth to train the registration network.

\subsection{Our baseline (Supervised Method)}
The registration network is also trained in a supervised manner as done in FlowNet \cite{dosovitskiy2015flownet} to verify the efficacy of the unsupervised method. More specifically, the deformation field produced by traditional registration algorithms is used as the ground truth to train the registration network. We regard the network trained by this supervised strategy as the baseline of our work.

\subsection{Implementation}
Our model is trained using Caffe \cite{jia2014caffe}. K40 GPU and CUDA 7.0 are used for training acceleration. We choose Adam \cite{kingma2014adam} as the optimization method with $\beta _{1}=0.9$ and $\beta_{2}=0.999$. The weight decay is 0.0005 and the batch size is 32. FlowNet is finetuned from the FlowNetSimple model pretrained on the Flying Chair dataset \cite{dosovitskiy2015flownet}.

For the unsupervised method, the learning rate is $10^{-5}$ when the training begins. We halve the learning rate after 10 epochs and keep training for another 7 epochs. The parameters $\alpha$ and $\beta$ introduced in Section \ref{loss} that balance different loss functions are set to 1 and 0.05 respectively for all datasets.

For the supervised method, the learning rate is $10^{-4}$  when the training begins. We halve the learning rate after 10 epochs and keep training for another 7 epochs.

\subsection{Parameters of Traditional Registration Algorithms}
The registration parameters used for the experimentation can be found at the source codes of ITK  \cite{ITK_code}. `DeformableRegistration $\bf{x}$.cxx' in this website corresponds to `itk $\bf{x}$' in Table \ref{table:results_brain_data_supervised} and \ref{table:results_liver_data_supervised}. For Elastix, BSpline is selected as the transformation model. The source code of Elastix used for registering MRI brain dataset can be found at \cite{Elastix_code_brain} and for registering  liver CT dataset can be found at \cite{Elastix_code_liver}. For ANTs, the following parameter settings are adopted to register MRI brain dataset:
\begin{Verbatim}[fontsize=\scriptsize]
antsRegistration -d 2 -m CC[${fixed},${moving},1,4]
               -t SyN[0.5] -c [600x600x50x10x0,0,5]
               -s 5x4x3x2x1vox -f 5x4x3x2x1 -u 1 -o ${prefix}
\end{Verbatim}
we adopt the following parameter settings to register liver CT dataset:
\begin{Verbatim}[fontsize=\scriptsize]
antsRegistration -d 2 -m CC[${fixed},${moving},1,4] 
               -t SyN[0.5] -c [600x600x50x10x0,0,5]
               -s 8x6x4x2x1vox -f 8x6x4x2x1 -u 1 -o ${prefix}
\end{Verbatim}

\subsection{Evaluation}
We use the same set of evaluation metrics for all datasets.
\subsubsection{Jaccard Coefficient}
The first metric is Jaccard coefficient that measures the overlap of ground truth segmentation masks. It is defined as $|A\cap B| / |A|\cup B|$ where A is the segmentation mask of the fixed image and B is the deformed segmentation mask of the moving image.
\subsubsection{Distance Between Corresponding Landmarks}
The second metric is introduced to measure the capacity of algorithms to register fine-grained structures. The registration error on a pair of images is quantified as the average 2D Euclidean distance between a landmark in the warped image and its corresponding landmark in the fixed image.

\subsection{Experiments on MRI Brain Registration}
\subsubsection{Dataset}
\label{dataset}
\begin{figure}[!htbp]
\centering
\includegraphics[width=60mm]{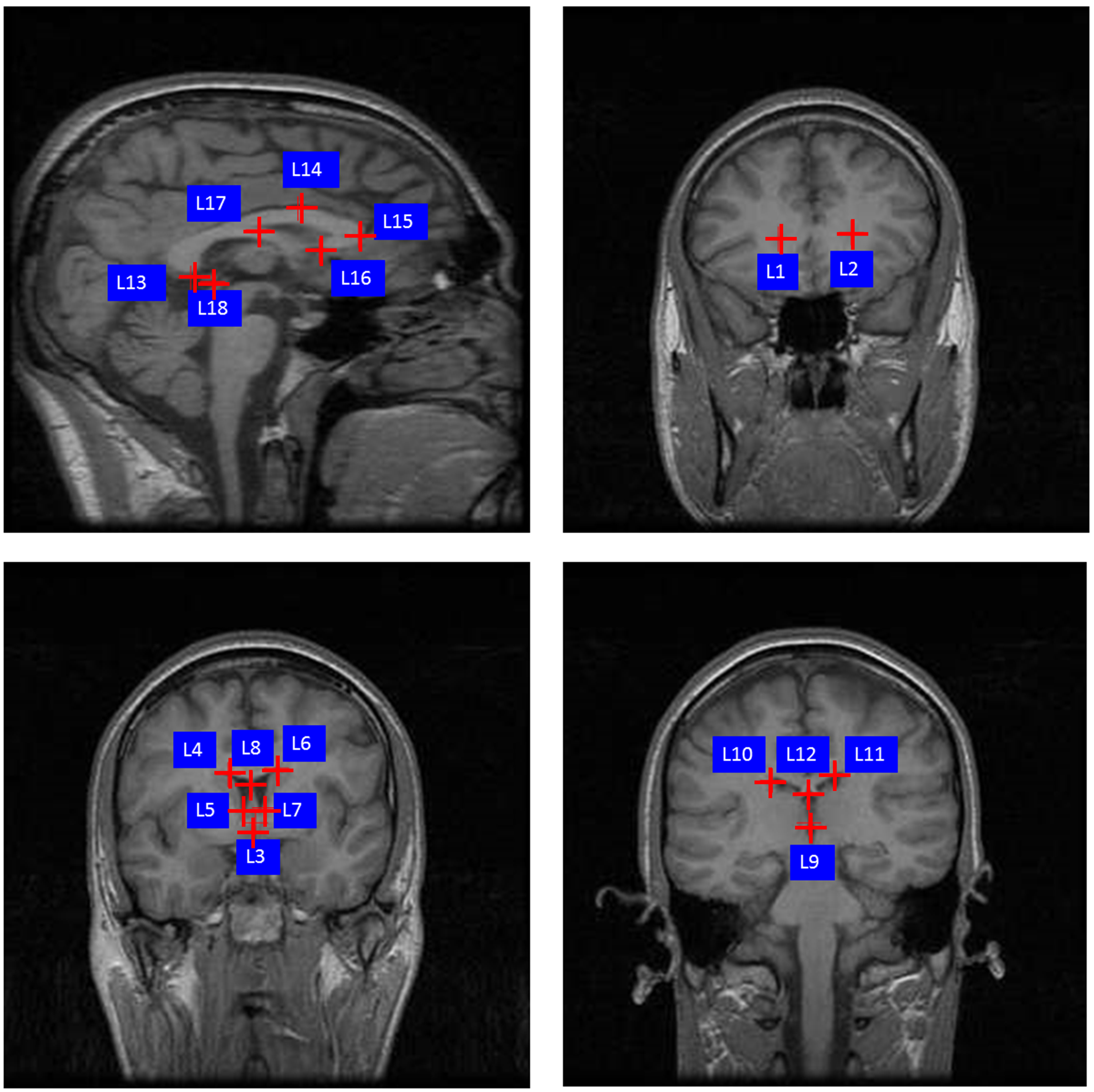}
\caption{This sketch illustrates the 18 landmarks selected in the brain dataset. L1: right lateral ventricle posterior, L2: left lateral ventricle posterior, L3: anterior commissure corresponds to the midpoint of decussation of the anterior commissure on the coronal AC plane, L4: right lateral ventricle superior, L5: right lateral ventricle inferior, L6: left lateral ventricle superior, L7: left lateral ventricle inferior, L8: middle of lateral ventricle, L9: posterior commissure corresponds to the midpoint of decussation, L10: right lateral ventricle superior, L11: left lateral ventricle superior, L12: middle of lateral ventricle, L13: corpus callosum inferior, L14: corpus callosum superior, L15: corpus callosum anterior, L16: corpus callosum posterior tip of genu corresponds to the location of the most posterior point of corpus callosum posterior tip of genu on the midsagittal planes, L17: corpus callosum fornix junction, L18: pineal body.}
\label{fig:brain_landmarks}
\end{figure}

The T1-weighted MRI brain data with ground-truth segmentation are selected from the LONI Probabilistic Brain Atlas (LPBA40) \cite{Shattuck2008Construction}, which consists of images from 40 subjects. We discard twenty images with tilted head positions and select the remaining twenty subjects. This dataset provides ground-truth segmentation masks. Eighteen well-defined anatomic landmarks (see Figure \ref{fig:brain_landmarks}) that are distributed mainly in the lateral ventricle and the median sagittal plane \cite{grachev1999method} are manually annotated by three doctors, and the average coordinates from three doctors are considered as the ground-truth positions of the landmarks. The original size of the 3D brain MRI volume is 256$\times$124$\times$256 voxels, which are zero-padded to 256$\times$128$\times$256 and resized to 256$\times$256$\times$256 voxels. Affine transformation (implemented by ANTs with mutual information as the metric) is applied to each 3D brain image before we slice 3D volume into 2D images. During the training phase, a pair of slicing planes A and B at the same position of the MRI volumes are interchangeably treated as a pair of fixed and moving images. This procedure produces a total of 291,840 (20$\times$19$\times$256$\times$3, 20$\times$19 subject pairs, 256 positions and 3 directions) 2D images for training and evaluation.

Additional MRI brain data is needed to improve the performance of our unsupervised methods. We randomly select 35 normal patients provided by ADNI \cite{adni} (Alzheimer's Disease Neuroimaging Initiative) to enlarge our training set. The original size of additional MRI brain data is 196$\times$256$\times$256 voxels, which is rotated into 256$\times$196$\times$256 and resized into 256$\times$256$\times$256 voxels. Then, the additional MRI brain data is preprocessed in the same way as the data of the LONI Probabilistic Brain Atlas (LPBA40) \cite{Shattuck2008Construction}.

\subsubsection{Experiment Results}
Fourfold cross validation is adopted in the experiment and the results are reported  on the 20,520 (20$\times$19$\times$18$\times$3) 2D slice pairs containing the same corresponding landmarks.

\begin{figure*}[!htbp]
\centering
\includegraphics[width=183mm]{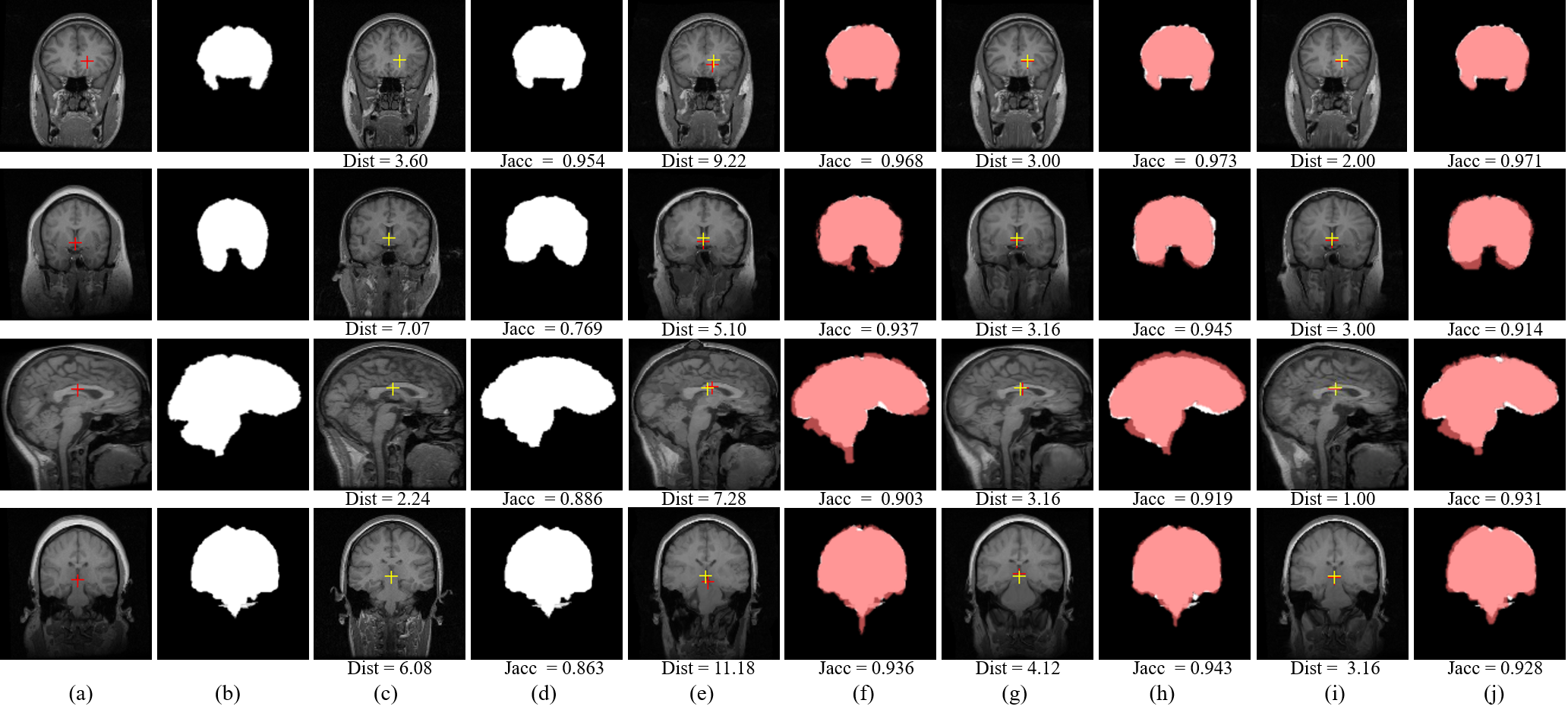}
\caption{Illustration of the brain registration performance of the proposed unsupervised methods, our best supervised baseline methods and the best traditional registration algorithms (itk16) with the best Dist: (a) Moving image, (b) Ground truth segmentation mask of moving image, (c) Fixed image, (d) Ground truth segmentation mask of fixed image. (e), (g), (i) respectively denote the moving images warped by the best traditional registration algorithm (itk16), our best supervised baseline method (itk16), our best unsupervised method (PE). The translucent red masks in (f), (h) and (j) respectively correspond to (e), (g), (i) and denote the warped ground truth segmentation mask of the moving images. The white masks in (f), (h), (j) are the ground-truth segmentation mask of the fixed image. The red and yellow crosses denote landmarks of moving image and fixed image, respectively. Dist in (c) and Jacc in (d) denote no registration.}
\label{fig:brain_result}
\end{figure*}

 Table \ref{table:results_brain_data}  quantitatively shows the performance of our unsupervised methods, our best baseline (supervised methods) and the best traditional registration algorithms. Jaccard Coefficient (Jacc) and Distance Between Corresponding Landmarks (Dist) are used as evaluation metrics. The running time (Rt) for each algorithm to register a pair of images is reported. The unsupervised methods PN and PE are 100x faster than traditional methods while achieving superior registration performance (Dist and Jacc). Besides, the unsupervised methods are superior to supervised methods. Figure \ref{fig:brain_result} illustrates the registration results of different methods.

\begin{table}[!htbp]
\renewcommand{\arraystretch}{1.2}
\caption{Performance of various methods with brain data. In our method, $<P, E, N>$ denote the inclusion of photometric loss $\mathcal{L}_{photometric}$, edge-aware smothness loss $\mathcal{L}_{smoothE}$ and normal smothness loss $\mathcal{L}_{smoothN}$, respectively. Then the size of training data is enlarged to improve the performance of unsupervised method. No registration means that no deformation occurs. Variations of various unsupervised methods proposed in the paper are usually different in the training but mostly share the
same architecture in testing; their run time speeds are therefore approximately the same.}
\label{table:results_brain_data}
\centering
\begin{tabular}{m{0.48\linewidth}<{\centering}|m{0.1\linewidth}<{\centering}|m{0.1\linewidth}<{\centering}|m{0.1\linewidth}<{\centering}}
\hline
Method & Dist & Jacc & Rt (s)\\
\hline
no registration & 4.04 & 0.908 & $\slash$\\
\hline
traditional method :&{}&{}&{}\\
traditional best (Dist) - itk16 & 3.21 & 0.948 &7.396\\
traditional best (Jacc) - ants & 3.33 & 0.955 &14.221\\
\hline
supervised (our baseline) :&{}&{}&{}\\
our best baseline (Dist) - itk16 & 3.14 & 0.956 &\bf{0.053}\\
our best baseline (Jacc) - itk16 & 3.14 & 0.956 &\bf{0.053}\\
\hline
unsupervised (our method):& {} & {} &{}\\
PN &3.06&0.953& \bf{0.053}\\
PE &3.05&0.951& \bf{0.053}\\
\hline
more unlabeled data (our method):&{} &{} &{}\\
PN &3.02&0.956&\bf{0.053}\\
PE &\bf{3.01}&\bf{0.958}&\bf{0.053}\\
\hline
\end{tabular}
\end{table}

\begin{table}[!htbp]
\renewcommand{\arraystretch}{1.2}
\caption{Performance of various traditional methods and our supervised baseline methods with brain data.}
\label{table:results_brain_data_supervised}
\centering
\begin{tabular}{m{0.25\linewidth}<{\centering}|m{0.065\linewidth}<{\centering}|m{0.065\linewidth}<{\centering}|m{0.065\linewidth}<{\centering} |m{0.065\linewidth}<{\centering}|m{0.065\linewidth}<{\centering}|m{0.065\linewidth}<{\centering}}
\hline
\multirow{2}{*}{} & \multicolumn{3}{c|}{Traditional method} & \multicolumn{3}{c}{Our baseline}\\ 
\hline
& Dist & Jacc & Rt (s)& Dist & Jacc & Rt (s)\\

\hline
itk1 (FEM)&3.40 & 0.948&25.860&4.54&0.897&\bf{0.053}\\
itk2 (Demons)& 3.33 & 0.934&5.426&3.34&0.934&\bf{0.053}\\
itk3 (Demons)& 3.22 & 0.936&10.428 &3.14&0.940&\bf{0.053}\\
itk5 (Demons) & 4.27 & 0.910&6.812&3.59&0.920&\bf{0.053}\\
itk13 (BSplines)& 3.22 & 0.938&50.205&3.26&0.936&\bf{0.053}\\
itk16 (Demons)& \bf{3.21} & 0.948&7.396&\bf{3.14}&\bf{0.956}&\bf{0.053}\\
itk17 (Demons)& 3.93 & 0.941&13.541&3.62 &0.940&\bf{0.053}\\
elastix& 3.45 & 0.951&25.840&3.43&0.950&\bf{0.053}\\
ants&3.33&\bf{0.955}&14.221&3.16&0.954&\bf{0.053}\\
\hline
\end{tabular}
\end{table}

Table \ref{table:results_brain_data_supervised} compares our supervised baseline methods to traditional registration algorithms, namely Insight Segmentation and Registration Toolkit (ITK) \cite{ITK}, Elastix \cite{klein2010elastix} and Advanced Normalization Tools (ANTs) \cite{avants2009advanced}. Note that our baseline models are trained in a supervised manner by regarding the results of the traditional algorithms in the same row of Table \ref{table:results_liver_data_supervised} as the ground truth. 

\subsection{Training with additional unlabeled data}
One advantage of unsupervised learning algorithms is that they do not require labeled training data. Therefore, the training set can be easily enlarged to further improve the registration performance without human labeling effort. As mentioned in \ref{dataset},  Additional MRI brain data of 35 patients provided by ADNI \cite{adni} (Alzheimer's Disease Neuroimaging Initiative) are selected to enlarge our training set. In the experiment, we retrain the unsupervised PN and PE model with this enlarged dataset and obtain an improvement in the registration performance. Notably, PE model achieves the best performance with Dist decreasing from 3.05 to 3.01 and Jacc increasing from 0.951 to 0.958, as is shown in Table \ref{table:results_brain_data}.

\subsection{Experiments on CT Liver Registration}
\subsubsection{Dataset}

\begin{figure}[!hbtp]
\centering
\includegraphics[width=60mm]{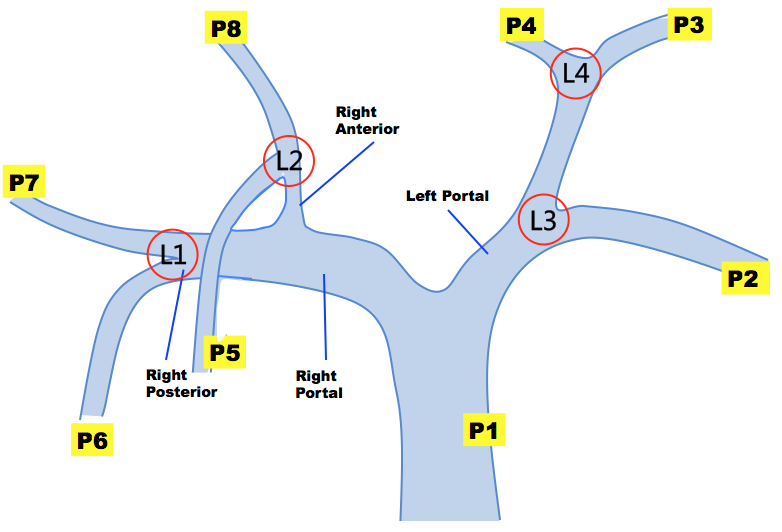}
\caption{This sketch illustrates the landmarks selected in the CT liver dataset. P1: hepatic portal,
P2: superior branch of left lobe, P3: inferior branch of left lobe, P4: medial branch of left lobe, P5 and P8: anterior branch of right, P6: inferior branch of right lobe, P7: superior branch of right lobe. L1, L2, L3, L4 are the selected landmarks, where L1 is the intersection point of P6 and P7, L2 is the intersection point of P5 and P8, L3 is the intersection point of P6 and P7 and L4 is the first bifurcation point of left portal.}
\label{fig:liver_landmarks}
\end{figure}

The 3D liver CT dataset is provided by the MICCAI 2007 Grand Challenge \cite{murphy2011evaluation}, which consists of images from 20 subjects. We discard two anomalous subjects and select the remaining 18 subjects. This dataset only provides ground-truth segmentation masks, and the coordinates of the landmarks are manually annotated by three doctors. Four landmarks (L1, L2, L3, L4) on the liver portal vein are selected (see Figure \ref{fig:liver_landmarks}). Each doctor labels the coordinates of the landmarks separately in 3D volumes via the ITK-SNAP tool. The average coordinates from the three doctors are considered the ground-truth positions of the landmarks. The regions of interest (128$\times$128$\times$128 voxels) containing the liver are extracted for further processing. As our network currently only supports 2D input images, we slice 3D CT volume along three orthogonal axes. Affine transformation (implemented by ANTs with mutual information as the metric) is applied to each 3D CT image before we slice 3D volume into 2D images. During the training phase, a pair of slicing planes A and B at the same position of the CT volumes are interchangeably treated as a pair of fixed and moving images. This procedure produces a total of 117,504 (18$\times$17$\times$3$\times$128, 18$\times$17 subject pairs, 3 directions and 128 positions) 2D images for training and evaluation.

Additional CT liver data provided by LiTS \cite{lits} (Liver Tumor Segmentation Challenge) from Training Batch 1 (it contains 3D CT volumes from 28 patients) are selected to enlarge our training set.  The regions of interest ( resized to 128$\times$128$\times$128 voxels) containing the liver are extracted for further processing. These data are preprocessed in the same way as the data of MICCAI 2007 Grand Challenge \cite{murphy2011evaluation}.

\subsubsection{Experiment Results}
\begin{figure*}[!tp]
\centering
\includegraphics[width=183mm]{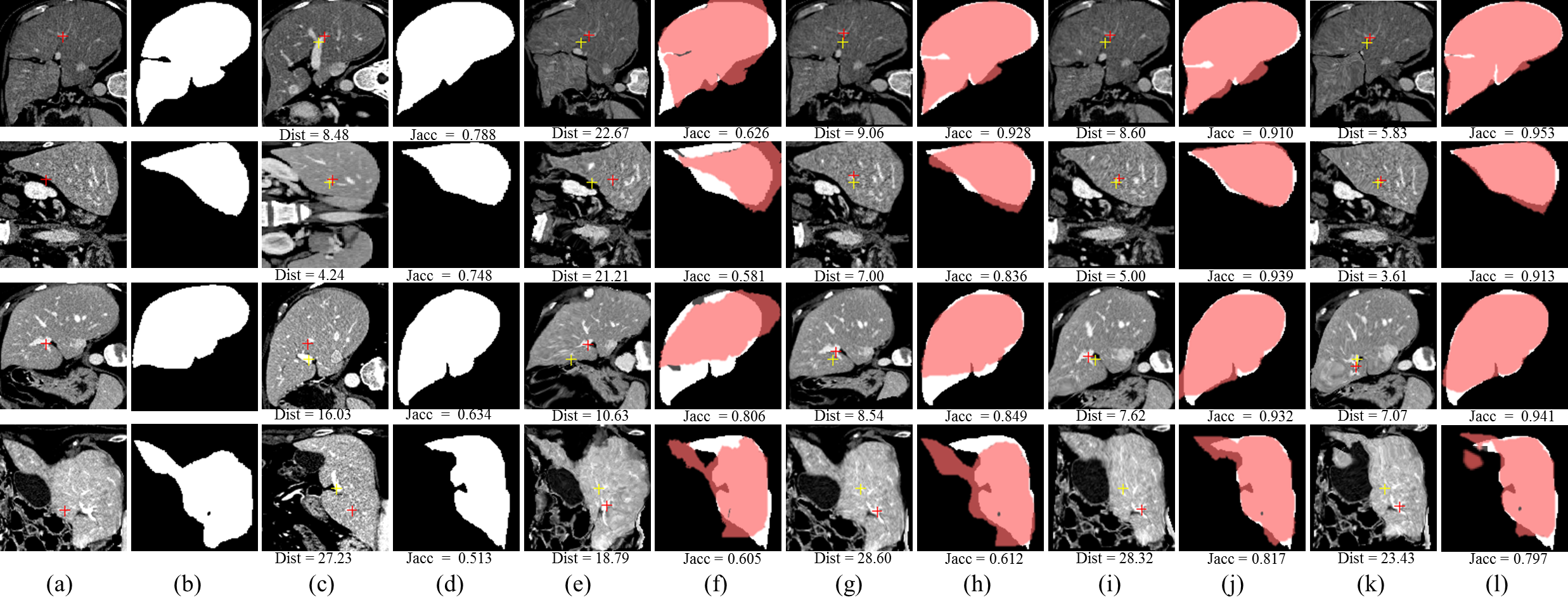}
\caption{Illustration of the liver registration performance of the proposed unsupervised methods, our best supervised baseline (w/ mask) methods and one the best traditional registration algorithms (Elastix) with the best Dist: (a) Moving image, (b) Ground truth segmentation mask of moving image, (c) Fixed image, (d) Ground truth segmentation mask of fixed image. (e), (g), (i) and (k) respectively denote the moving images warped by the best traditional registration algorithm (elastix), our best supervised baseline method (itk16), our best unsupervised method PN. The translucent red masks in (f), (h), (j), (l) respectively correspond to (e), (g), (i), (k) and denote the warped ground truth segmentation mask of the moving images. The white masks in (f), (h), (j), (l) are the ground-truth segmentation mask of the fixed image. The red and yellow crosses denote landmarks of moving image and fixed image, respectively. Dist in (c) and Jacc in (d) denote no registration.}
\label{fig:liver_result}
\end{figure*}

Trifold cross validation is adopted in the experiment and the results are reported only on the 3,672 (18$\times$17$\times$4$\times$3$\times$1) 2D slice pairs containing the same corresponding landmarks. Table  \ref{table:results_liver_data} quantitatively shows the performance of our unsupervised methods, our best baseline (supervised methods) and the best traditional registration algorithms. We retrain the unsupervised PN and  PE model with this enlarged dataset and obtain an improvement on the registration performance. Notably,  Dist of PE model decreases from 13.79 to 13.54 and Jacc increases from 0.837 to 0.845, as is shown in Table \ref{table:results_liver_data}. Figure \ref {fig:liver_result} illustrates the registration results of different methods.

\begin{table}[!hbtp]
\renewcommand{\arraystretch}{1.2}
\caption{Performance of various methods with liver data. In our method, $<P, E, N>$ denote the inclusion of photometric loss $\mathcal{L}_{photometric}$, edge-aware smothness loss $\mathcal{L}_{smoothE}$ and normal smothness loss $\mathcal{L}_{smoothN}$, respectively.Then the size of training data is enlarged to improve the performance of unsupervised method. No registration means that no deformation occurs. Variations of various unsupervised methods proposed in the paper are usually different in the training but mostly share the
same architecture in testing; their run time speeds are therefore approximately the same.}
\label{table:results_liver_data}
\centering
\begin{tabular}{m{0.48\linewidth}<{\centering}|m{0.1\linewidth}<{\centering}|m{0.1\linewidth}<{\centering}|m{0.1\linewidth}<{\centering}}
\hline
Method & Dist & Jacc & Rt (s)\\
\hline
no registration & 13.46 & 0.665& $\slash$ \\
\hline
traditional method :&{}&{}&{}\\
traditional best (Dist) - elastix& \bf{11.77} & 0.817& 29.935\\
traditional best (Jacc) - itk17& 12.68 & 0.836& 6.886\\
\hline
supervised (our baseline) :&{}&{}&{}\\
our best baseline (Dist)  - itk16 & 11.78 & 0.780 &\bf{0.032}\\
our best baseline (Jacc)  - ants & 12.09 & 0.791 &\bf{0.032}\\
\hline
unsupervised (our method):& {} & {} &{}\\
PN &12.51&0.822& \bf{0.032}\\
PE &13.79&0.837& \bf{0.032}\\
\hline
more unlabeled data (our method):&{}&{}&{}\\
PN &12.35&0.831&\bf{0.032}\\
PE &13.54&\bf{0.845}&\bf{0.032}\\
\hline
\end{tabular}
\end{table}

\begin{table}[!hbtp]
\renewcommand{\arraystretch}{1.2}
\caption{Performance of various traditional methods and our supervised baseline methods with liver data.}
\label{table:results_liver_data_supervised}
\centering
\begin{tabular}{m{0.25\linewidth}<{\centering}|m{0.065\linewidth}<{\centering}|m{0.065\linewidth}<{\centering}|m{0.065\linewidth}<{\centering}|m{0.065\linewidth}<{\centering}|m{0.065\linewidth}<{\centering}|m{0.065\linewidth}<{\centering}}
\hline
\multirow{2}{*}{} & \multicolumn{3}{c|}{Traditional method} & \multicolumn{3}{c}{Our baseline}\\ 
\hline 
& Dist & Jacc & Rt (s)& Dist & Jacc & Rt (s)\\

\hline
itk1 (FEM)&12.84 & 0.729 &11.014&13.13&0.677&\bf{0.032}\\
itk2 (Demons)& 13.39 & 0.704 & 2.954&13.39&0.688&\bf{0.032}\\
itk3 (Demons)& 13.28 & 0.772 & 5.103&12.82 &0.753&\bf{0.032}\\
itk5 (Demons) & 13.90 & 0.785 & 4.312&13.35 &0.705&\bf{0.032}\\
itk13 (BSplines)& 12.95 & 0.725 & 15.540&13.24&0.705&\bf{0.032}\\
itk16 (Demons)& 11.94 & 0.807 & 8.809& \bf{11.78}&0.780&\bf{0.032}\\
itk17 (Demons)& 12.68& \bf{0.836} & 6.886& 12.79& 0.781&\bf{0.032}\\
elastix& \bf{11.77 }& 0.817 &30.602&12.73 &0.790 &\bf{0.032}\\
ants&12.51&0.822&5.131&12.09&\bf{0.791}&\bf{0.032}\\
\hline
\end{tabular}
\end{table}

Table \ref{table:results_liver_data_supervised} compares our baselines with several traditional registration algorithms. Our baselines run faster than traditional registration algorithms and achieve superior performance both in terms of Dist and Jacc.

Although our methods outperform traditional methods, the performance of liver registration is  still far from clinical requirements. In order to enhance the accuracy of registration, we introduce ROI segmentation mask  into our unsupervised pipeline as an extra component. The details of ROI segmentation mask are illustrated in supplement material.


\section{Conclusion}
In this paper, we have developed an end-to-end framework using unsupervised fully convolutional neural networks to perform medical image registration. The proposed network is trained in an unsupervised manner without any ground-truth deformation.  Experiments demonstrate that our methods achieve the state-of-the-art results on the MRI brain dataset in both accuracy and registration speed, and achieve the comparable results on the CT liver dataset. We achieve a 100x speed-up compared to traditional image registration methods. The scope of our proposed methods is quite broad and can be widely applied to various medical image registration and computer vision applications.


%


\section*{Supplement Materials }
\label{supp}

\begin{figure}[!htbp]
\centering
\includegraphics[width=90mm]{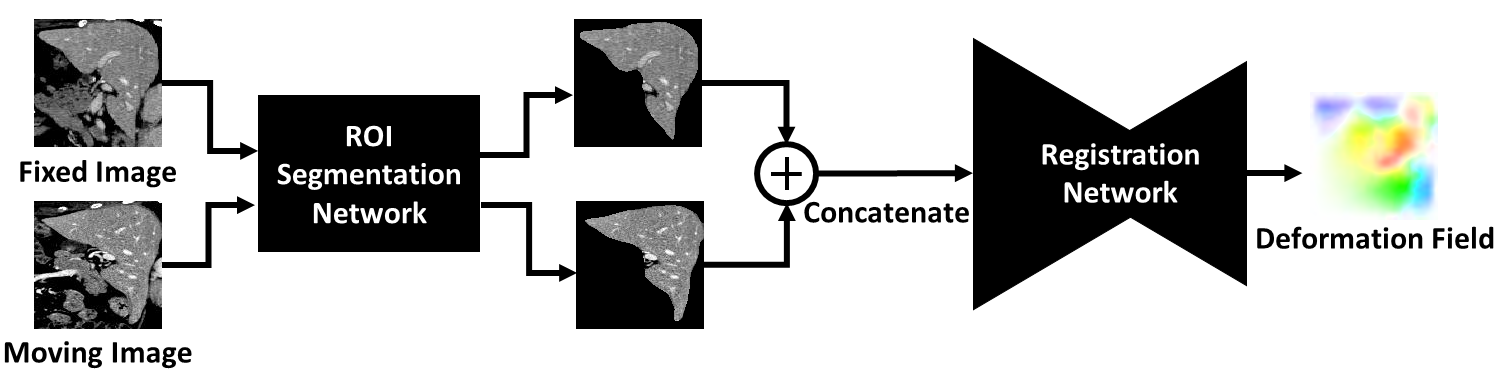}
\caption{This illustrates a brief structure of our fully convolutional image-to-image registration network. The ROI segmentation network predicts the ROI segmentation masks of the input fixed and moving images to reduce background noises and reserve the ROI. The registration network takes two images and outputs a deformation field.}
\label{fig:framework_simple}
\end{figure}

Compared with brain images, the background of liver images is more complicated as there are various anatomical structures in the abdomen. The background noise, such as other visceral organs or vessel,  has an extremely negative impact on liver registration. For such tasks, ROI segmentation is a prerequisite for successful image registration as shown in \cite{murphy2011evaluation,greene2009constrained,ferrante2017slice}, as it can focus the registration process on specific regions of interest (ROI) and avoid the undesired alignment of artifacts. 
We also introduce ROI segmentation mask module, which reduces background noise and interference. That is, we first segment tissues to be registrated. After that, we registrate the tissues based on the segmentation masks. Our system pipeline with ROI segmentation mask module is shown in Figure \ref{fig:framework_simple}.

Our motivation to introduce ROI segmentation mask is two-fold. First, our unsupervised flow networks under an image-to-image paradigm allow the ROI mask to be conveniently implemented through back-propagation. Second, having ROI mask theoretically and mathematically greatly enhances registration performance \cite{murphy2011evaluation}; this is evident in our experiments where a significant performance boost is observed using the module. Our ROI segmentation mask module is performed under a Holistically-nested nets (HNN) \cite{xie2015holistically,lee2015deeply} paradigm, as it can perform multi-scale and multi-level learning under deep supervision within FCNs. HNN is finetuned from the pretrained 5-stage VGG \cite{simonyan2014very}. The learning rate of HNN is $10^{-4}$  when the training begins for quick mask initialization. After 4 epochs, the learning rate of the HNN part decreases to $10^{-7}$.

We find that using photometric difference loss alone results in poor quality deformation field in some extreme cases where there are serious illumination changes. To overcome these problems, the ROI segmentation mask produced by HNN is leveraged to guide the deformation field learning process. The predicted segmentation mask is found to be both accurate and robust to illumination changes, which motivates us to add a ROI boundary overlapping loss $\mathcal{L}_{overlap}^{s}$ to our system,
\begin{equation}
\mathcal{L}_{overlap}^{s} =  \sum_{\mathbf{x}\in \Omega}||\tilde{D}^{s}\left ( \mathbf{x} \right ) - D_{F}^{s}(\mathbf{x})|| ,
\end{equation} 
where $\tilde{D}^{s}\left ( \mathbf{x} \right )= D_{M}^{s}\left ( \mathbf{x}+\mathbf{u^{s}(x)} \right )$ denotes the ROI segmentation mask of the moving image resized to scale $s$ and warped by STN according to deformation field $\mathbf{u^{s}(x)}$, $D_{F}^{s}(\mathbf{x})$ is the ROI segmentation mask of the fixed image resized to scale $s$.


\subsection{Experiments on CT Liver Registration}
\subsubsection{Experiments Result}
As is illustrated in Table \ref{table:results_liver_data}, the performance of liver registration is beyond satisfying, which is far from clinical requirements. Thus, we add another extra experiment by introducing ROI segmentation module into the image-to-image deformable regsitration module. To demonstrate the contribution of the ROI segmentation module, we compare the performance of different models with  the ROI segmentation module, denoted by w/ mask. Table \ref{table:results_liver_data_mask} quantitatively shows the performance of our unsupervised methods with mask, our best baseline (supervised methods) and the best traditional registration algorithms with mask.

The unsupervised methods PMN w/ mask and PME w/ mask exhibit capability to align object boundaries by achieving high Jaccard Coefficient 0.903 and 0.905 respectively. And it's evidenced that ROI segmentation mask has significantly increased the performance of liver registration both in Dist and Jacc. Notably, methods based on convolution neural networks achieve 100x speedup compared to traditional methods. Figure \ref{fig:liver_result} illustrates the registration results of different methods on liver CT data.

 Table \ref{table:results_liver_data_mask_supervised} shows the performance of various traditional methods and our supervised baseline methods with mask.



\begin{table}[!htbp]
\renewcommand{\arraystretch}{1.2}
\caption{Performance of various methods with mask using liver data. In our method, $<P, M, E, N>$ denote the inclusion of photometric loss $\mathcal{L}_{photometric}$, ROI boundary overlapping loss $\mathcal{L}_{overlap}$, edge-aware smothness loss $\mathcal{L}_{smoothE}$ and normal smothness loss $\mathcal{L}_{smoothN}$, respectively. Then the size of training data is enlarged to improve the performance of unsupervised method with mask. Variations of various unsupervised methods proposed in the paper are usually different in the training but mostly share the same architecture in testing; their run time speeds are therefore approximately the same.}
\label{table:results_liver_data_mask}
\centering
\begin{tabular}{m{0.48\linewidth}<{\centering}|m{0.1\linewidth}<{\centering}|m{0.1\linewidth}<{\centering}|m{0.1\linewidth}<{\centering}}
\hline
Method & Dist & Jacc & Rt (s)\\
\hline
trditional method: w/mask :&{}&{}&{}\\
traditional best (Dist)  - itk16 & \bf{11.00} & 0.934 &3.274\\
traditional best (Jacc)  - itk17 & 11.25 & 0.981 &8.690\\
\hline
supervised w/mask (our baseline) :&{}&{}&{}\\
our best baseline (Dist)  - itk16 & 11.49 & 0.868 &\bf{0.057}\\
our best baseline (Jacc)  - itk16 & 11.49 & 0.868 &\bf{0.057}\\
\hline
unsupervised w/ mask (our method):& {} & {} &{}\\
PN &12.14&0.872& \bf{0.057}\\
PE &13.63&0.872&\bf{ 0.057}\\
PMN &11.74&0.903&\bf{ 0.057}\\
PME &12.87&0.905& \bf{0.057}\\
\hline
more unlabeled data w/ mask (our method):&{} &{} &{}\\
PMN &11.21&0.917&\bf{ 0.057}\\
PME &12.54&\bf{0.918}&\bf{0.057}\\

\hline
\end{tabular}
\end{table}


\begin{table}[!hbtp]
\renewcommand{\arraystretch}{1.2}
\caption{Performance of various traditional methods and our supervised and unsupervised methods with mask using liver data.}
\label{table:results_liver_data_mask_supervised}
\centering
\begin{tabular}{m{0.25\linewidth}<{\centering}|m{0.065\linewidth}<{\centering}|m{0.065\linewidth}<{\centering}|m{0.065\linewidth}<{\centering}|m{0.065\linewidth}<{\centering}|m{0.065\linewidth}<{\centering}|m{0.065\linewidth}<{\centering}}
\hline
\multirow{2}{*}{} & \multicolumn{3}{c|}{Traditional method /w} & \multicolumn{3}{c}{Our baseline /w}\\ 
\hline 
& Dist & Jacc & Rt (s)& Dist & Jacc & Rt (s)\\
\hline
itk1 (FEM)&12.82 & 0.734 &11.748&12.92&0.682&0.057\\
itk2 (Demons)& 13.35 & 0.709 & 2.793&13.38&0.687&0.057\\
itk3 (Demons)& 13.21 & 0.865 & 4.985&13.26&0.762&0.057\\
itk5 (Demons) & 13.49 & 0.948 & 4.555 &12.82 &0.803&0.057\\
itk13 (BSplines)& 12.56 & 0.783 & 8.390&12.04&0.755&0.057\\
itk16 (Demons)& \bf{11.00 }& 0.934 & 3.274&\bf{11.49}&\bf{0.868}&0.057\\
itk17 (Demons)& 11.25& \bf{0.981} & 8.690&11.87 &0.845&0.057\\
elastix& 11.26 & 0.967 &30.543&11.53&0.847&0.057\\
ants&11.68&0.967&3.710&11.52&0.846&0.057\\
\hline
\end{tabular}
\end{table}


\subsubsection{Training with additional unlabeled data}
Given that ROI segmentation model is able to boost the performance of liver registration, in the experiment  we retrain the PMN w/ mask model and PME  w/ mask model with this enlarged dataset and observe an improvement on the registration performance, with Dist decreasing from 11.74 to 11.21 and Jacc increasing from 0.903 to 0.917, as shown in Table \ref{table:results_liver_data_mask}. Noted that we do not use any label during this processing, neither the liver segmentation mask  provided by LiTS nor optical flow label generated by traditional methods. The ROI segmentation modules are trained only by original data of MICCAI 2007 Grand Challenge \cite{murphy2011evaluation}, and the registration modules are trained in a completely unsupervised manner.

\begin{figure*}[!tp]
\centering
\includegraphics[width=183mm]{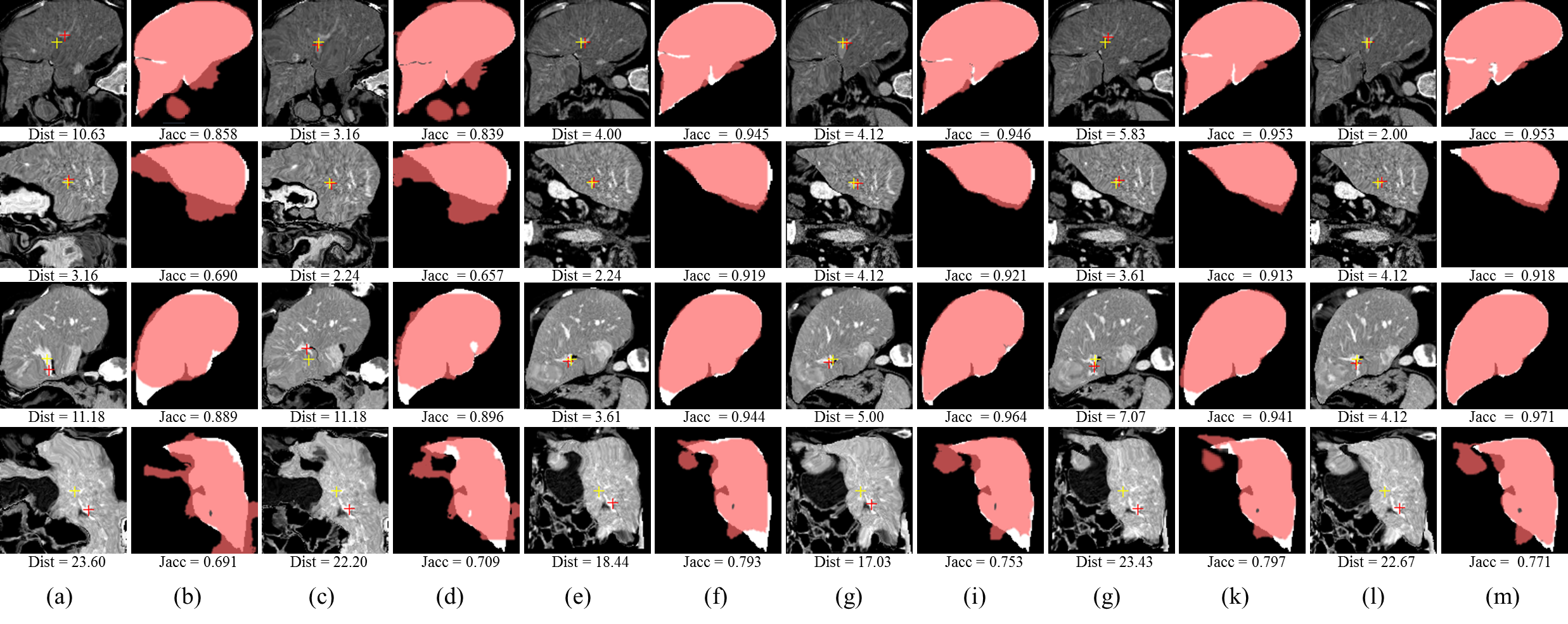}
\caption{Illustration of the liver registration performance of the proposed unsupervised methods with and without mask: (a), (c), (e), (g), (i) and (k) respectively denote the moving images warped by PN, PE, PE with more unlabeled data, PMN w/ mask and PME w/ mask, PME w/ mask with more unlabeled data. The translucent red masks in (b), (d), (f), (h), (j), (l) respectively correspond to (a), (c), (e), (g), (i), (k) and denote the warped ground truth segmentation mask of the moving images. The white masks in (b), (d), (f), (h), (j), (l) are the ground-truth segmentation mask of the fixed image. The red and yellow crosses denote landmarks of moving image and fixed image, respectively.}
\label{fig:liver_result_unsp}
\end{figure*}

Figure \ref{fig:liver_result_unsp} illustrates the registration results of unsupervised methods  w/ mask (in Table \ref{table:results_liver_data_mask}) and without mask (in Table \ref{table:results_liver_data}).  

\subsection{Experiments on MRI brain Registration}
We also validate the effect of ROI segmentation module on brain MRI dataset, as is shown in Table \ref{table:results_brain_data_mask} and Table \ref{table:results_brain_data_mask_supervised}. We retrain the unsupervised PN w/ mask model and PE w/ mask model with additional dataset provided by ADNI \cite{adni} (Alzheimer's Disease Neuroimaging Initiative, and observe an improvement in the registration performance. Note that the ROI segmentation modules  are trained only by original data provided the  LONI Probabilistic Brain Atlas (LPBA40) \cite{Shattuck2008Construction}. The result demonstrates that the ROI segmentation module can improve the performance of brain registration limitedly compared to liver. It is due to that the background noise of brain images is much lower than liver images.

\begin{figure*}[!tp]
\centering
\includegraphics[width=183mm]{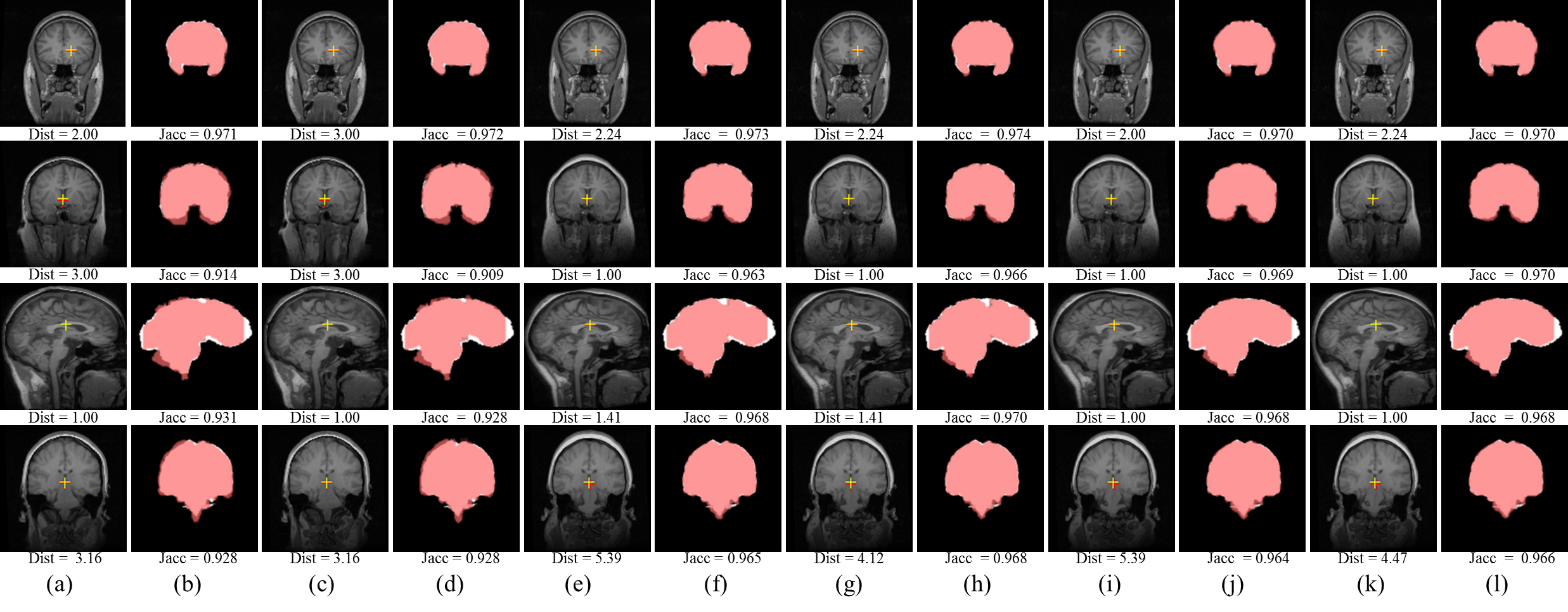}
\caption{Illustration of the brain registration performance of the proposed unsupervised methods with and without mask: (a), (c), (e) , (g), (i) and  (k) respectively denote the moving images warped by PN, PE, PE with more unlabeled data, PN w/ mask and PE w/ mask, PE w/ mask with more unlabeled data. The translucent red masks in (b), (d), (f), (h), (j), (l) respectively correspond to (a), (c), (e), (g), (i) and (k) denote the warped ground truth segmentation mask of the moving images. The white masks in (b), (d), (f), (h), (j), (l) are the ground-truth segmentation mask of the fixed image. The red and yellow crosses denote landmarks of moving image and fixed image, respectively.}
\label{fig:brain_result_unsp}
\end{figure*}

Figure \ref{fig:brain_result_unsp} illustrates the registration results of unsupervised methods  w/ mask (in Table \ref{table:results_brain_data_mask}) and without mask (in Table \ref{table:results_brain_data}).  


\begin{table}[!htbp]
\renewcommand{\arraystretch}{1.2}
\caption{Performance of various methods with mask using brain data. In our method, $<P, E, N>$ denote the inclusion of photometric loss $\mathcal{L}_{photometric}$, edge-aware smothness loss $\mathcal{L}_{smoothE}$ and normal smothness loss $\mathcal{L}_{smoothN}$, respectively. Then the size of training data is enlarged to improve the performance of unsupervised method. Variations of various unsupervised methods proposed in the paper are usually different in the training but mostly share the same architecture in testing; their run time speeds are therefore approximately the same.}
\label{table:results_brain_data_mask}
\centering
\begin{tabular}{m{0.48\linewidth}<{\centering}|m{0.1\linewidth}<{\centering}|m{0.1\linewidth}<{\centering}|m{0.1\linewidth}<{\centering}}
\hline
Method & Dist & Jacc & Rt (s)\\
\hline
traditional method w/ mask:&{}&{}&{}\\
traditional best (Dist) - itk3& 3.14 & 0.998 &10.123\\
traditional best (Jacc) - itk17 & 3.54 & \bf{1.000 }&12.450\\
\hline
supervised (our baseline) :&{}&{}&{}\\
our best baseline (Dist) - itk3 & \bf{3.04} & 0.964 &\bf{0.094}\\
our best baseline (Jacc) - itk16 & 3.18 & 0.967 &\bf{0.094}\\
\hline
unsupervised w/ mask (our method):& {} & {} &{}\\
PN &3.37 &0.971 & \bf{0.094}\\
PE &3.35&0.951& \bf{0.094}\\
\hline
more unlabeled data w/ mask (our method):&{} &{} &{}\\
PN &3.32&0.974&\bf{0.094}\\
PE &3.31&0.955&\bf{0.094}\\
\hline
\end{tabular}
\end{table}

\begin{table}[!hbtp]
\renewcommand{\arraystretch}{1.2}
\caption{Performance of various traditional methods and our supervised and unsupervised methods with mask using brain data.}
\label{table:results_brain_data_mask_supervised}
\centering
\begin{tabular}{m{0.25\linewidth}<{\centering}|m{0.065\linewidth}<{\centering}|m{0.065\linewidth}<{\centering}|m{0.065\linewidth}<{\centering}|m{0.065\linewidth}<{\centering}|m{0.065\linewidth}<{\centering}|m{0.065\linewidth}<{\centering}}
\hline
\multirow{2}{*}{} & \multicolumn{3}{c|}{Traditional method /w} & \multicolumn{3}{c}{Our baseline /w}\\ 
\hline 
& Dist & Jacc & Rt (s)& Dist & Jacc & Rt (s)\\

\hline
itk1 (FEM)&3.44&0.973&26.740&3.36&0.954&\bf{0.094}\\
itk2 (Demons)&3.26&0.958&5.412&3.30&0.943&\bf{0.094}\\
itk3 (Demons)&\bf{3.14}&0.998&10.123&\bf{3.04} &0.964&\bf{0.094}\\
itk5 (Demons) &4.31&\bf{1.000}&6.654&3.45&0.939&\bf{0.094}\\
itk13 (BSplines)&\bf{3.14}&0.968&51.739&3.15&0.960&\bf{0.094}\\
itk16 (Demons)&3.18&0.998&5.717&3.18&\bf{0.967}&\bf{0.094}\\
itk17 (Demons)&3.53&\bf{1.000}&12.450&3.43 &0.951&\bf{0.094}\\
elastix &3.38&0.988&25.981&3.23&0.955&\bf{0.094}\\
ants &3.57&0.994&7.767&3.24&0.966&\bf{0.094}\\
\hline
\end{tabular}
\end{table}

\section*{Acknowledgment}

The authors would like to thank all the dataset providers for making their databases publicly available.

\ifCLASSOPTIONcaptionsoff
  \newpage
\fi



%

\bibliographystyle{IEEEtran}
\bibliography{IEEEabrv,./ref}

\begin{thebibliography}{10}
\providecommand{\url}[1]{#1}
\csname url@samestyle\endcsname
\providecommand{\newblock}{\relax}
\providecommand{\bibinfo}[2]{#2}
\providecommand{\BIBentrySTDinterwordspacing}{\spaceskip=0pt\relax}
\providecommand{\BIBentryALTinterwordstretchfactor}{4}
\providecommand{\BIBentryALTinterwordspacing}{\spaceskip=\fontdimen2\font plus
\BIBentryALTinterwordstretchfactor\fontdimen3\font minus
  \fontdimen4\font\relax}
\providecommand{\BIBforeignlanguage}[2]{{%
\expandafter\ifx\csname l@#1\endcsname\relax
\typeout{** WARNING: IEEEtran.bst: No hyphenation pattern has been}%
\typeout{** loaded for the language `#1'. Using the pattern for}%
\typeout{** the default language instead.}%
\else
\language=\csname l@#1\endcsname
\fi
#2}}
\providecommand{\BIBdecl}{\relax}
\BIBdecl

\bibitem{ibanez2005itk}
L.~Ibanez, W.~Schroeder, L.~Ng, and J.~Cates, ``The itk software guide,'' 2005.

\bibitem{avants2010ants}
B.~B. Avants, N.~J. Tustison, G.~Song, and J.~C. Gee, ``Ants, advanced
  normalization tools,'' \url{http://www.picsl.upenn.edu/ANTS/}, 2010.

\bibitem{klein2010elastix}
S.~Klein, M.~Staring, K.~Murphy, M.~A. Viergever, and J.~P. Pluim, ``Elastix: a
  toolbox for intensity-based medical image registration,'' \emph{TMI},
  vol.~29, no.~1, pp. 196--205, 2010.

\bibitem{ashburner2007fast}
J.~Ashburner, ``A fast diffeomorphic image registration algorithm,''
  \emph{Neuroimage}, vol.~38, no.~1, pp. 95--113, 2007.

\bibitem{vercauteren2009diffeomorphic}
T.~Vercauteren, X.~Pennec, A.~Perchant, and N.~Ayache, ``Diffeomorphic demons:
  Efficient non-parametric image registration,'' \emph{NeuroImage}, vol.~45,
  no.~1, pp. S61--S72, 2009.

\bibitem{song2010lung}
G.~Song, N.~Tustison, B.~Avants, and J.~C. Gee, ``Lung ct image registration
  using diffeomorphic transformation models,'' \emph{Medical image analysis for
  the clinic: a grand challenge}, pp. 23--32, 2010.

\bibitem{thirion1998image}
J.-P. Thirion, ``Image matching as a diffusion process: an analogy with
  maxwell's demons,'' \emph{Medical image analysis}, vol.~2, no.~3, pp.
  243--260, 1998.

\bibitem{xu20163d}
Y.~Xu, C.~Xu, X.~Kuang, H.~Wang, E.~I. Chang, W.~Huang, Y.~Fan \emph{et~al.},
  ``3d-sift-flow for atlas-based ct liver image segmentation,'' \emph{Medical
  physics}, vol.~43, no.~5, pp. 2229--2241, 2016.

\bibitem{guetter2005learning}
C.~Guetter, C.~Xu, F.~Sauer, and J.~Hornegger, ``Learning based non-rigid
  multi-modal image registration using kullback-leibler divergence,''
  \emph{MICCAI}, pp. 255--262, 2005.

\bibitem{krizhevsky2012imagenet}
A.~Krizhevsky, I.~Sutskever, and G.~E. Hinton, ``Imagenet classification with
  deep convolutional neural networks,'' in \emph{NIPS}, 2012, pp. 1097--1105.

\bibitem{simonyan2014very}
K.~Simonyan and A.~Zisserman, ``Very deep convolutional networks for
  large-scale image recognition,'' \emph{arXiv preprint arXiv:1409.1556}, 2014.

\bibitem{ren2015faster}
S.~Ren, K.~He, R.~Girshick, and J.~Sun, ``Faster r-cnn: Towards real-time
  object detection with region proposal networks,'' in \emph{NIPS}, 2015, pp.
  91--99.

\bibitem{long2015fully}
J.~Long, E.~Shelhamer, and T.~Darrell, ``Fully convolutional networks for
  semantic segmentation,'' in \emph{CVPR}, 2015, pp. 3431--3440.

\bibitem{xie2015holistically}
S.~Xie and Z.~Tu, ``Holistically-nested edge detection,'' in \emph{ICCV}, 2015,
  pp. 1395--1403.

\bibitem{zbontar2016stereo}
J.~Zbontar and Y.~LeCun, ``Stereo matching by training a convolutional neural
  network to compare image patches,'' \emph{JMLR}, vol.~17, no. 1-32, p.~2,
  2016.

\bibitem{wei2016dense}
L.~Wei, Q.~Huang, D.~Ceylan, E.~Vouga, and H.~Li, ``Dense human body
  correspondences using convolutional networks,'' in \emph{CVPR}, 2016, pp.
  1544--1553.

\bibitem{litjens2017survey}
G.~Litjens, T.~Kooi, B.~E. Bejnordi, A.~A.~A. Setio, F.~Ciompi, M.~Ghafoorian,
  J.~A. van~der Laak, B.~van Ginneken, and C.~I. S{\'a}nchez, ``A survey on
  deep learning in medical image analysis,'' \emph{arXiv preprint
  arXiv:1702.05747}, 2017.

\bibitem{yang2016fast}
X.~Yang, R.~Kwitt, and M.~Niethammer, ``Fast predictive image registration,''
  in \emph{LABELS}, 2016, pp. 48--57.

\bibitem{wu2013unsupervised}
G.~Wu, M.~Kim, Q.~Wang, Y.~Gao, S.~Liao, and D.~Shen, ``Unsupervised deep
  feature learning for deformable registration of mr brain images,'' in
  \emph{MICCAI}, 2013, pp. 649--656.

\bibitem{miao2016cnn}
S.~Miao, Z.~J. Wang, and R.~Liao, ``A cnn regression approach for real-time
  2d/3d registration,'' \emph{TMI}, vol.~35, no.~5, pp. 1352--1363, 2016.

\bibitem{fischer2015flownet}
A.~Dosovitskiy, P.~Fischer, E.~Ilg, P.~Hausser, C.~Hazirbas, V.~Golkov,
  P.~van~der Smagt, D.~Cremers, and T.~Brox, ``Flownet: Learning optical flow
  with convolutional networks,'' in \emph{ICCV}, 2015, pp. 2758--2766.

\bibitem{jaderberg2015spatial}
M.~Jaderberg, K.~Simonyan, A.~Zisserman \emph{et~al.}, ``Spatial transformer
  networks,'' in \emph{NIPS}, 2015, pp. 2017--2025.

\bibitem{ren2017unsupervised}
Z.~Ren, J.~Yan, B.~Ni, B.~Liu, X.~Yang, and H.~Zha, ``Unsupervised deep
  learning for optical flow estimation,'' in \emph{AAAI}, 2017.

\bibitem{yu2016back}
J.~Y. Jason, A.~W. Harley, and K.~G. Derpanis, ``Back to basics: Unsupervised
  learning of optical flow via brightness constancy and motion smoothness,'' in
  \emph{ECCV}, 2016, pp. 3--10.

\bibitem{garg2016unsupervised}
R.~Garg, G.~Carneiro, and I.~Reid, ``Unsupervised cnn for single view depth
  estimation: Geometry to the rescue,'' in \emph{ECCV}, 2016, pp. 740--756.

\bibitem{godard2016unsupervised}
C.~Godard, O.~Mac~Aodha, and G.~J. Brostow, ``Unsupervised monocular depth
  estimation with left-right consistency,'' in \emph{CVPR}, 2017.

\bibitem{kanazawa2016warpnet}
A.~Kanazawa, D.~W. Jacobs, and M.~Chandraker, ``Warpnet: Weakly supervised
  matching for single-view reconstruction,'' in \emph{CVPR}, 2016, pp.
  3253--3261.

\bibitem{dosovitskiy2015flownet}
A.~Dosovitskiy, P.~Fischer, E.~Ilg, P.~Hausser, C.~Hazirbas, V.~Golkov,
  P.~van~der Smagt, D.~Cremers, and T.~Brox, ``Flownet: Learning optical flow
  with convolutional networks,'' in \emph{ICCV}, 2015, pp. 2758--2766.

\bibitem{kybic2003fast}
J.~Kybic and M.~Unser, ``Fast parametric elastic image registration,''
  \emph{IEEE transactions on image processing}, vol.~12, no.~11, pp.
  1427--1442, 2003.

\bibitem{viola1997alignment}
P.~Viola and W.~M. Wells~III, ``Alignment by maximization of mutual
  information,'' \emph{International journal of computer vision}, vol.~24,
  no.~2, pp. 137--154, 1997.

\bibitem{rohlfing2003volume}
T.~Rohlfing, C.~R. Maurer, D.~A. Bluemke, and M.~A. Jacobs, ``Volume-preserving
  nonrigid registration of mr breast images using free-form deformation with an
  incompressibility constraint,'' \emph{TMI}, vol.~22, no.~6, pp. 730--741,
  2003.

\bibitem{penney1998comparison}
G.~P. Penney, J.~Weese, J.~A. Little, P.~Desmedt, D.~L. Hill \emph{et~al.}, ``A
  comparison of similarity measures for use in 2-d-3-d medical image
  registration,'' \emph{TMI}, vol.~17, no.~4, pp. 586--595, 1998.

\bibitem{staring2007rigidity}
M.~Staring, S.~Klein, and J.~P. Pluim, ``A rigidity penalty term for nonrigid
  registration,'' \emph{Medical physics}, vol.~34, no.~11, pp. 4098--4108,
  2007.

\bibitem{lester1999survey}
H.~Lester and S.~R. Arridge, ``A survey of hierarchical non-linear medical
  image registration,'' \emph{Pattern recognition}, vol.~32, no.~1, pp.
  129--149, 1999.

\bibitem{rueckert1999nonrigid}
D.~Rueckert, L.~I. Sonoda, C.~Hayes, D.~L. Hill, M.~O. Leach, and D.~J. Hawkes,
  ``Nonrigid registration using free-form deformations: application to breast
  mr images,'' \emph{TMI}, vol.~18, no.~8, pp. 712--721, 1999.

\bibitem{Sokooti:2017}
{Hessam Sokooti and Bob de Vos and Floris Berendsen and Boudewijn P.F.
  Lelieveldt and Ivana Išgum and Marius Staring}, ``{Nonrigid Image
  Registration Using Multi-Scale 3D Convolutional Neural Networks},'' in
  \emph{{MICCAI}}, {Quebec,Canada}, {2017}.

\bibitem{heise2013pm}
P.~Heise, S.~Klose, B.~Jensen, and A.~Knoll, ``Pm-huber: Patchmatch with huber
  regularization for stereo matching,'' in \emph{ICCV}, 2013, pp. 2360--2367.

\bibitem{avants2009advanced}
B.~B. Avants, N.~Tustison, and G.~Song, ``Advanced normalization tools
  (ants),'' \emph{Insight j}, vol.~2, pp. 1--35, 2009.

\bibitem{jia2014caffe}
Y.~Jia, E.~Shelhamer, J.~Donahue, S.~Karayev, J.~Long, R.~Girshick,
  S.~Guadarrama, and T.~Darrell, ``Caffe: Convolutional architecture for fast
  feature embedding,'' in \emph{Proceedings of the 22nd ACM international
  conference on Multimedia}, 2014, pp. 675--678.

\bibitem{kingma2014adam}
D.~Kingma and J.~Ba, ``Adam: A method for stochastic optimization,''
  \emph{arXiv preprint arXiv:1412.6980}, 2014.

\bibitem{ITK_code}
``Source codes of itk,''
  \url{https://github.com/InsightSoftwareConsortium/ITK/tree/master/Examples/RegistrationITKv4}.

\bibitem{Elastix_code_brain}
``Source codes of elastix,''
  \url{http://elastix.bigr.nl/wiki/images/d/d5/Par0035.Hammers.MI.bs.1.ASGDPrime.txt}.

\bibitem{Elastix_code_liver}
``Source codes of elastix,''
  \url{http://elastix.bigr.nl/wiki/images/c/c9/Parameters_BSpline.txt}.

\bibitem{Shattuck2008Construction}
D.~W. Shattuck, M.~Mirza, V.~Adisetiyo, C.~Hojatkashani, G.~Salamon, K.~L.
  Narr, R.~A. Poldrack, R.~M. Bilder, and A.~W. Toga, ``Construction of a 3d
  probabilistic atlas of human cortical structures,'' \emph{Neuroimage},
  vol.~39, no.~3, pp. 1064--1080, 2008.

\bibitem{grachev1999method}
I.~D. Grachev, D.~Berdichevsky, S.~L. Rauch, S.~Heckers, D.~N. Kennedy, V.~S.
  Caviness, and N.~M. Alpert, ``A method for assessing the accuracy of
  intersubject registration of the human brain using anatomic landmarks,''
  \emph{Neuroimage}, vol.~9, no.~2, pp. 250--268, 1999.

\bibitem{adni}
``The alzheimer's disease neuroimaging initiative (adni),''
  \url{http://adni.loni.usc.edu/methods/mri-analysis/adni-standardized-data}.

\bibitem{ITK}
``Insight segmentation and registration toolkit (itk),''
  \url{https://itk.org/}.

\bibitem{murphy2011evaluation}
K.~Murphy, B.~Van~Ginneken, J.~M. Reinhardt, S.~Kabus, K.~Ding, X.~Deng,
  K.~Cao, K.~Du, G.~E. Christensen, V.~Garcia \emph{et~al.}, ``Evaluation of
  registration methods on thoracic ct: the empire10 challenge,'' \emph{TMI},
  vol.~30, no.~11, pp. 1901--1920, 2011.

\bibitem{lits}
``Lits (liver tumor segmentation challenge),''
  \url{https://competitions.codalab.org/competitions/17094#learn_the_details-overview}.

\bibitem{greene2009constrained}
W.~H. Greene, S.~Chelikani, K.~Purushothaman, J.~Knisely, Z.~Chen,
  X.~Papademetris, L.~H. Staib, and J.~S. Duncan, ``Constrained non-rigid
  registration for use in image-guided adaptive radiotherapy,'' \emph{Medical
  image analysis}, vol.~13, no.~5, pp. 809--817, 2009.

\bibitem{ferrante2017slice}
E.~Ferrante and N.~Paragios, ``Slice-to-volume medical image registration: A
  survey,'' \emph{Medical Image Analysis}, vol.~39, pp. 101--123, 2017.

\bibitem{lee2015deeply}
C.-Y. Lee, S.~Xie, P.~Gallagher, Z.~Zhang, and Z.~Tu, ``Deeply-supervised
  nets,'' in \emph{Artificial Intelligence and Statistics}, 2015, pp. 562--570.

\end{thebibliography}
%

\end{document}